\documentclass[applsci,article,submit,pdftex,moreauthors]{Definitions/mdpi} 

\firstpage{1} 
\pubvolume{1}
\issuenum{1}
\articlenumber{0}
\pubyear{2023}
\copyrightyear{2023}
\datereceived{ } 
\daterevised{ } % Comment out if no revised date
\dateaccepted{ } 
\datepublished{ } 
\hreflink{https://doi.org/} % If needed use \linebreak
\usepackage{subcaption}
\usepackage{paralist}

\usepackage{metawriting/metawriting}
\usepackage{metawriting/sciabbr}
\ours{TWNet}

\Title{Tamed Warping Network for High-Resolution Semantic Video Segmentation}

\TitleCitation{TWNet}

\Author{Songyuan Li$^{1}$, Junyi Feng$^{1}$ and Xi Li$^{1}$*}
\AuthorNames{Songyuan Li, Junyi Feng and Xi Li}

\AuthorCitation{Li, S.; Feng, J.; Li, X.}
 \address{
  $^{1}$ \quad Zhejiang University}
\corres{Correspondence: xilizju@zju.edu.cn}

\firstnote{These authors contributed equally to this work.  } 
\abstract{Recent approaches for fast semantic video segmentation have reduced redundancy by warping feature maps across adjacent frames, greatly speeding up the inference phase. However, the accuracy drops seriously owing to the errors incurred by warping.
	In this paper, we propose a novel framework and design a simple and effective correction stage after warping. Specifically, we build a non-key-frame CNN, fusing warped context features with current spatial details. Based on the feature fusion, our Context Feature Rectification~(CFR) module learns the model's difference from a per-frame model to correct the warped features. Furthermore, our Residual-Guided Attention~(RGA) module utilizes the residual maps in the compressed domain to help CRF focus on error-prone regions. Results on Cityscapes show that the accuracy significantly increases from $67.3\%$ to $71.6\%$, and the speed edges down from $65.5$ FPS to $61.8$ FPS at a resolution of $1024\times 2048$. For non-rigid categories, \eg, ``human'' and ``object'', the improvements are even higher than 18 percentage points.
}

\keyword{Semantic Video Segmentation, Warping} 

\begin{document}
\section{Introduction}\label{sec:introduction}

%As an important and challenging problem in computer vision,
Semantic video segmentation aims to predict pixel-wise class labels for each frame in a video. 
A real-time solution to this task is challenging due to the stringent requirements of speed and space. Prevailing fast methods can be grouped into two major categories: per-frame and warping-based.
%In general, there are two kinds of approaches.
Per-frame methods decompose the \emph{video} task into a stream of mutually independent \emph{image} segmentation, usually reducing input resolution~\cite{segnet, dfanet, bisenet, BiAttnNet, EACNet,zhang2022laanet,zhang2022lightweight,hu2022joint,fan2022mlfnet} or adopting a lightweight CNN \cite{mobilenet, dfanet, df1-seg, swiftnet, mobilenetv2, bisenet, cas, icnet,liu2020efficient,xiao2022real,hu2020temporally} to meet the speed demand. In light of the visual continuity between adjacent video frames, warping-based methods~\cite{gadde2017semantic, jain2018fast, low-latency, dvsnet, dff} employ inter-frame motion estimation \cite{dvsnet, dff, jain2018fast} to reduce redundancy. %, utilizing the results from the previous frame to the current one. 

\begin{figure}[t]
    %\centering 

    \begin{adjustwidth}{-\extralength}{0cm}
        \begin{subfigure}{0.5\linewidth}
            \centering
            % not found
            \includegraphics[width=0.8\linewidth]{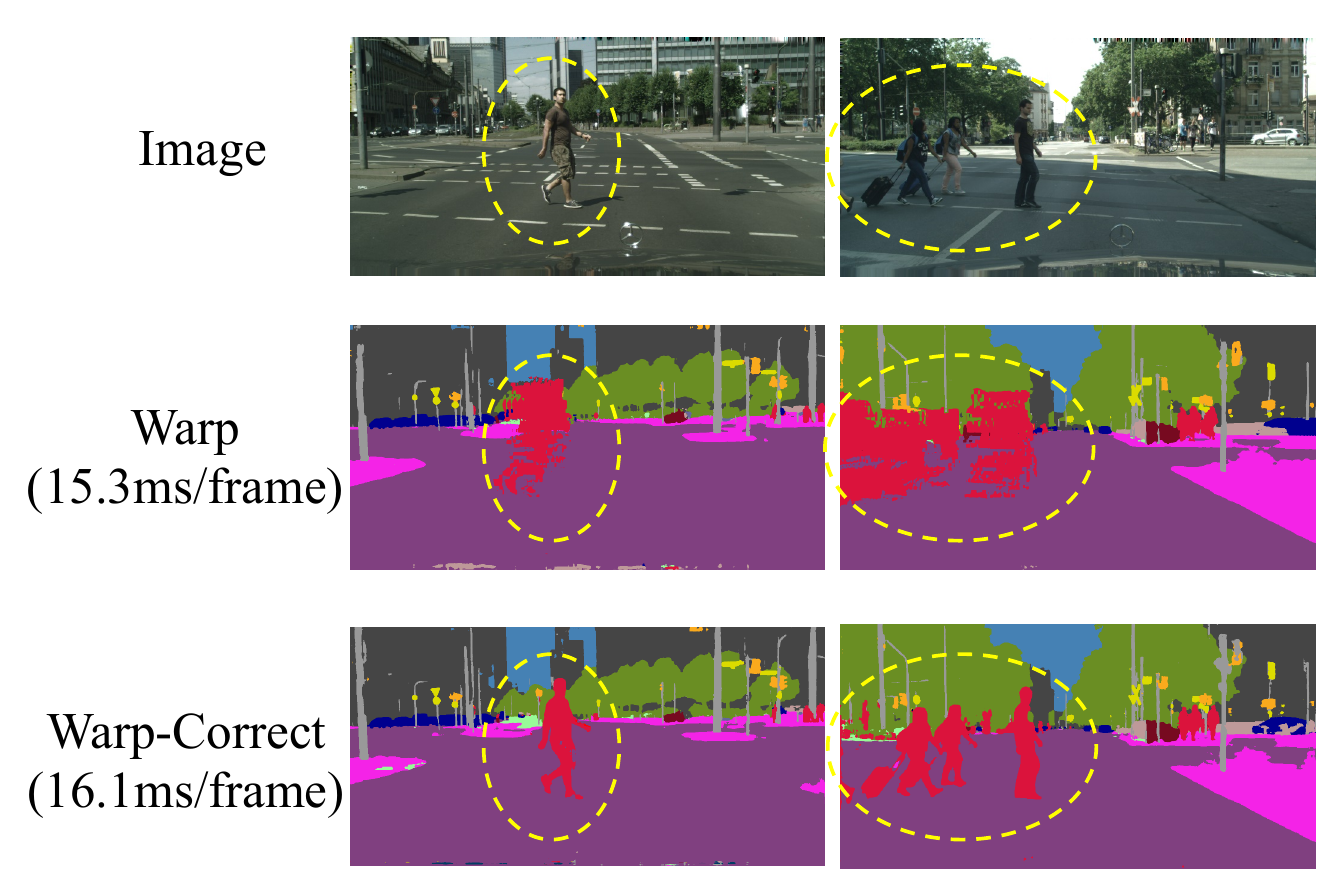}\\
            \caption{Walking people} \label{sfig:walking-people}

        \end{subfigure}%
        \begin{subfigure}{0.5\linewidth}
            \centering
            % not found
            \includegraphics[width=0.8\linewidth]{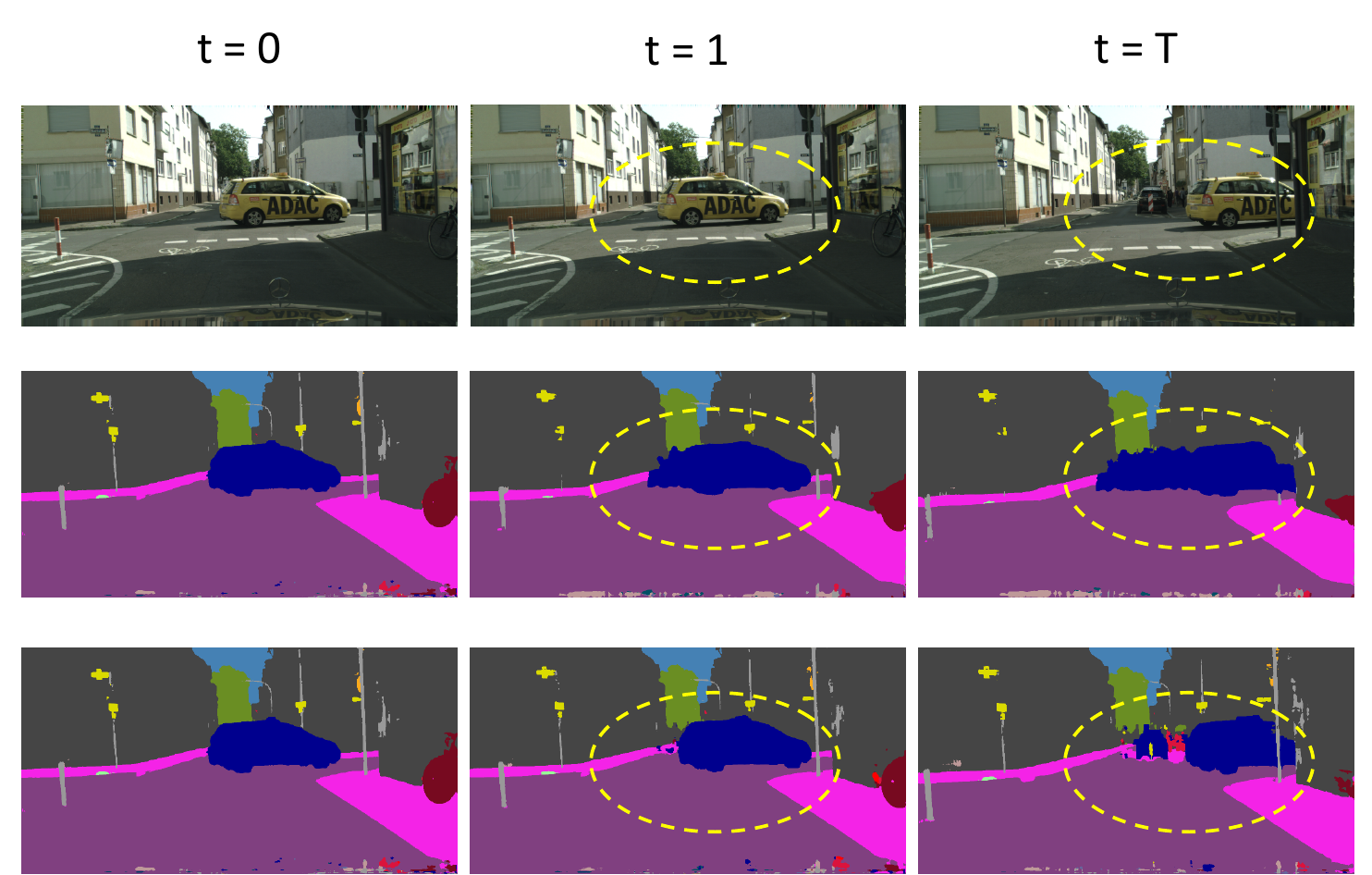}\\
            \caption{A moving car} \label{sfig:moving-car}
        \end{subfigure}%
    \centering
    \end{adjustwidth}
	\caption{Results of warping and warping with correction. (\subref{sfig:walking-people}) The walking people's limbs can be occluded by their own bodies a few frames ago but appear later, thus becoming severely deformed in later warped frames. (\subref{sfig:moving-car}) A moving car occludes distant objects which cannot be warped from previous frames.  By adding a correction stage, errors can be significantly alleviated as shown in the third row.} \label{fig:warping-problem}
\end{figure}

In spite of  boosting the inference speed, the accuracy of warping-based methods drops considerably due to warping itself. Warping inevitably introduces errors, which accumulate along consecutive non-key frames, making the later results almost unusable (\cref{fig:warping-problem}). Warping turns out to behave like a runaway fierce creature; the key to the issue is to tame it --- to take advantage of its acceleration and to keep it under control. To this end, we build a framework called Tamed Warping Network and add a correction stage after warping, utilizing spatial features and residuals to correct warped features.

% First, we propose the non-key-frame CNN~(NKFC) based on a U-shape FCN \cite{fpn, unet} to perform segmentation for non-key frames. 
% It has long been observed that high-level features change more slowly than the low-level ones \cite{clockwork, accel}. 
% Motivated by this, we make NKFC fuse the warped deep features from the previous frame with the features of the current frame from its shallow layers, as shown in Fig.~\ref{fig:warping_strategy}, making itself fast and able to retain spatial details. 
% With the help of the spatial details, we introduce a correction stage consisting of two modules. We design the context feature rectification~(CFR) module to learn the difference of the warping model from a per-frame model to correct the warped features. To help CFR focus on error-prone regions, we design the residual-guided attention~(RGA) module to utilize the \textit{Residual Maps} in the compressed domain.  

The contributions are summarized as follows:
 \begin{inparaenum}[1)]
 	\item 
	We propose a fast high-resolution semantic video segmentation framework utilizing compressed-domain motion vectors to do warping and fusing warped context features with current spatial details for the following feature correction. % with feature correction adapted to key frames and non-key frames to improve the accuracy of warping-based models.   % with feature correction to improve the accuracy of warping-based segmentation. % by introducing a lightweight correction stage.
	\item To alleviate the errors incurred by warping, we propose two correction modules, Context Feature Rectification~(CFR) and Residual-Guided Attention~(RGA). CFR fuses warped context features with the current spatial details to learn feature space residuals under the guidance of a per-frame model. To step further, RGA utilizes compressed-domain residuals to correct features learned from CFR. %  making the prediction for non-key frames accurate and robust. 
    \item
    Experiments % and CamVid
     show that TWNet is generic to backbone choices and significantly increases the mIoU of the baseline from $67.3\%$ to $71.6\%$. For non-rigid categories, the improvements are even higher than 18 percentage points.
 \end{inparaenum}

\section{Related Work}
\label{sec:related}
%	Driven by the development of deep convolutional neural networks~(DCNNs), 

%Real-time methods usually reduce the scale of input images or adopt lightweight networks (either a lightweight backbone or a customized lightweight architecture). In this work, we %utilize temporal continuity in video by warping and minimize its resulting errors by feature fusion. Next, we briefly summarize these two aspects.
	%In this section, 
% (either a light-weight CNN backbone, \eg, ResNet18~\cite{resnet} and MobileNet~\cite{mobilenet, mobilenetv2}, or a custom lightweight architecture~\cite{df1-seg, cas})
\subsection{Feature Fusion in Semantic Segmentation}
	% Segmentation
Recently, both high-quality models~\cite{deeplabv3+, ccnet, cfnet, pspnet, ann, danet, dynamic-filter,sharma2020ssfnet} %, such as PSPNet~\cite{pspnet}, DeeplabV3+~\cite{deeplabv3+}, and CFNet~\cite{cfnet}
and high-speed ones~\cite{dfanet, swiftnet, bisenet, icnet} %, such as ICNet~\cite{icnet}, BiSeNet~\cite{bisenet}, DFANet~\cite{dfanet} and SwiftNet~\cite{swiftnet}, 
show the importance of feature fusion from different layers. %~(or scales). 
Generally, shallow layers contain more low-level \textit{spatial} details, while deep layers more \textit{contextual} information. The combination of features from different layers improves the accuracy.
	
	% Relation
	Feature fusion in our proposed non-key-frame CNN (NKFC) is similar to those in \cite{fpn, unet,lin2020refineu}, where lateral connections are used to fuse the low-level~(spatial) and high-level~(context) features. 
%	The main difference between these two methods and 
In comparison, our NKFC only retains a few layers of the encoder to extracts low-level features and obtains high-level features by feature warping. % , while high-level features are obtained by feature warping. %warping the context features from the previous frame.
Thus, NKFC saves the heavy computations of context feature extraction.
	
	% Features in videos
	\vspace{1em}
\subsection{Warping-Based Video Segmentation}
	%In general, videos are of temporal continuity, i.e., consecutive video frames look similar,  
	%making it possible to perform inter-frame prediction. The process of mapping pixels from the previous frame to the current one according to a pixel-level motion map is called image warping~(Fig.~\ref{fig:inter-frame}). 
	%Each point in the motion map is a two-dimensional vector, $(\Delta x, \Delta y)$, representing the movement from the previous frame to the current one. Motion vectors and optical flows are two kinds of commonly used motion maps. In general, motion vectors, already contained in videos, are less precise than optical flows~(e.g., TV-L1~\cite{tv-l1} FlowNet2.0~\cite{flownetv2} and PWC-Net~\cite{pwcnet}). However, it takes extra time to perform optical flow estimation. 

	% warping
	Researchers have proposed many warping-based approaches~\cite{gadde2017semantic, jain2018fast, dvsnet, dff}. Some works adopt warping as a temporal constraint to enhance features for the sake of accuracy~\mbox{\cite{gadde2017semantic,liu2020efficient,hu2020temporally}}.
	%Gadde~et al.~\cite{gadde2017semantic} proposed to enhance the features of the current frame by introducing features from previous frames.
	For acceleration, Zhu~et al.~\mbox{\cite{dff}}, Xu~et al.~\mbox{\cite{dvsnet}} and Jain~et al.~\mbox{\cite{jain2018fast}} proposed to use feature warping to speed up their models. They divided frames into two types: key frames and non-key frames. Key frames are sent to the CNN for segmentation, while non-key frame results are obtained by warping. Recently, Hu~et al.~\cite{hu2020temporally} proposed to approximate high-level features by composing features from several shallower layers.
		% problems 1. bias, 2. acumulate  equation
		% bias, key-frame selection
		These approaches speed up the inference phase since the computational cost of warping is much less than that of CNN. However, both the accuracy and robustness of these methods deteriorate due to the following reasons. First, neither optical flows nor motion vectors can estimate the precise motion of all pixels. There always exist unavoidable biases~(errors) between the warped features and the expected ones. Second, in the case of consecutive non-key frames, cumulative errors lead to unusable results.
		To address error accumulation, Li~et al.~\cite{low-latency} and Xu~et al.~\cite{dvsnet} proposed to adaptively select key frames by predicting the confidence score for each frame. Jain~et al.~\cite{jain2018fast} introduced bi-directional features warping to improve accuracy. However, all these approaches lack the ability to correct warped features.
		
		% Ours
		%The main difference between the proposed Warp-Refine Network and previous warping-based methods is that our approach firstly includes the refinement stage after feature warping.

		% Our approach is related to motion compensation in two folds
			% warping
			% refine <-> residual in 
		%Our WRNet is related to motion compensation in the following aspects. First, the feature warping is similar with warping for images. Second, inspired by the correction step in motion compensation, we propose the context feature refinement~(CFR) module to perform correction in feature space. Last, observing the spatial consistency between image space and feature space, we propose to use the residual map in image space to guide the learning of CFR.

\section{Warping and Correction in Video Codecs}
\label{sec:prelim}
\begin{figure}[tb]

\begin{adjustwidth}{-\extralength}{0cm}
\begin{center}
   \includegraphics[width=0.8\linewidth]{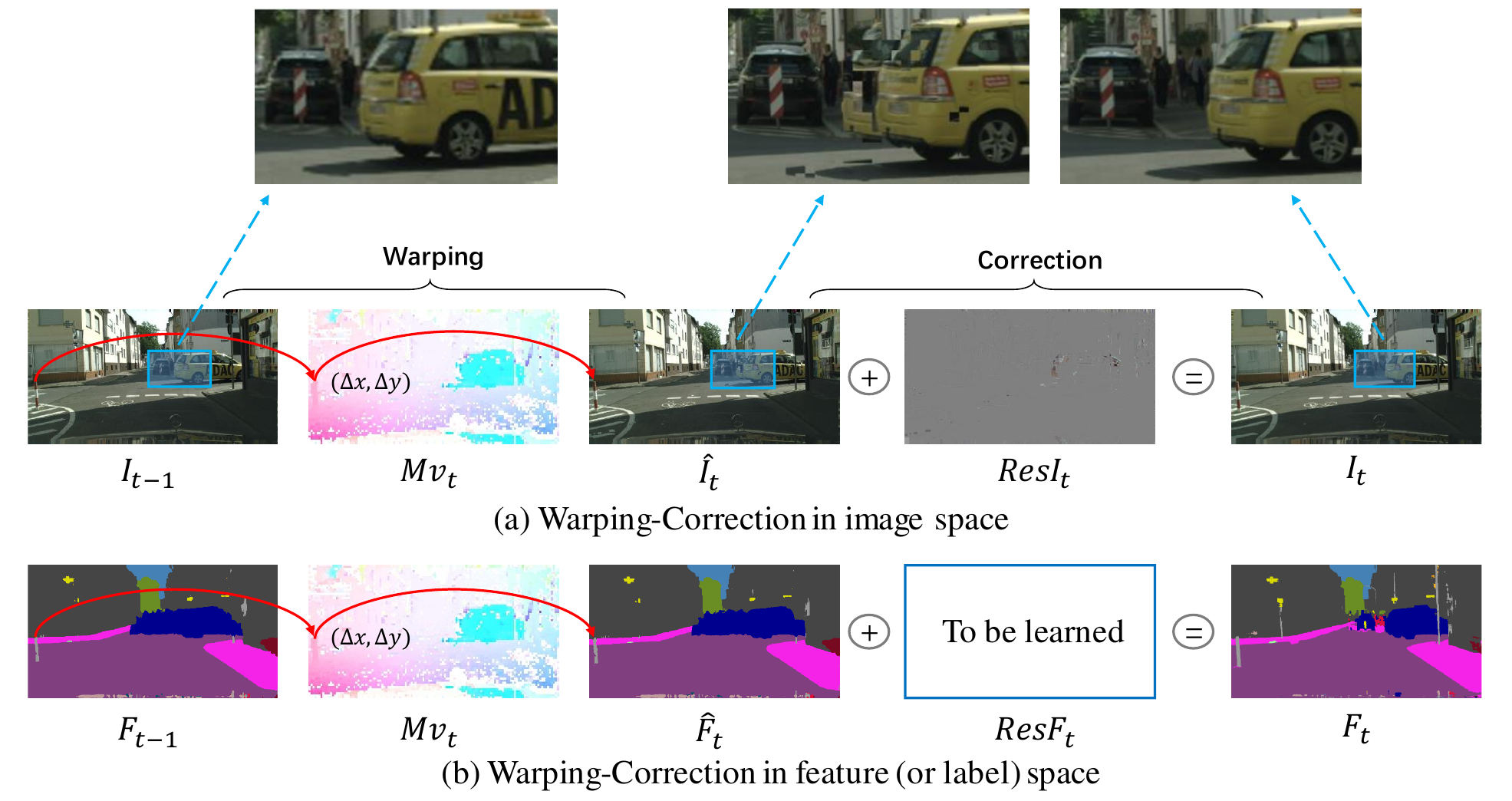}
\end{center}
\end{adjustwidth}
	\vspace{-0.8em}
   \caption{Illustration of \textbf{Warping} and \textbf{Correction}. (a) Video codecs warp the previous frame to the current one, and then add the residual map. (b) We propose to learn the \textbf{residual in feature space} to rectify the warped features. %For visualization, segmentation labels are used to represent features. 
   }
\label{fig:inter-frame}
\end{figure}

	%In this section, we describe the basics and the pipeline of warping-correction in video codecs.
	% warping
	Warping is an efficient operation for frame estimation. Given the previous frame $I_{t-1}$ and the motion vectors of the current frame $Mv_t$, the current frame $\hat{I}_t$ can be estimated by 
\begin{equation}
\begin{split}
	\hat{I}_t = \mathrm{warp}(I_{t-1}, Mv_t).
\end{split}
\label{eq:warp}			
\end{equation}
%
% For a particular location index $p\in \mathcal{Z}^{H\times W}$ in an image, the mapping function is given by
	% %
% \begin{equation}
% \begin{split}
	% \hat{I}_{t}[p] &= I_{t-1}[p-Mv_{t}[p]].
% \end{split}
% \label{eq:feature-warp}			
% \end{equation}
	% Residual
	However, there always exist biases between the warped image and the real one. %, especially in some complicated scenes and for the non-rigid moving objects. 
	 Modern video codecs %~(e.g., MPEG~\cite{mpeg}, H.264~\cite{h264}) 
	 add a correction step after image warping. (\cref{fig:inter-frame}(a)). 
	Specifically, the codec performs pixel-wise addition between the warped image $\hat{I}_t$ and the \textit{residual map} $ResI_t$. Each point in $ResI_t$ is a three-dimensional vector, $(\Delta r, \Delta g, \Delta b)$, which describes the color differences between the warped pixel and the expected one. The overall inter-frame prediction process is described as
		%the inter-frame prediction~(\ie, motion compensation) usually contains two steps. First, the codec performs image warping using motion vectors. After that, pixel-wise addition between the warped image~(may contain error) and the residual map is performed for correction. Let $X_t$ denotes the $t^{th}$ frame. $Mv_t$, $Res_t$ are the corresponding motion vector and residual, the prediction process is described as
\begin{equation}
\begin{split}
	I_t &= \mathrm{warp}(I_{t-1}, Mv_t) + ResI_t.
\end{split}
\label{eq:warp-refine}			
\end{equation}
%
	%The key difference between Eq.~\ref{eq:warp} and Eq.~\ref{eq:warp-refine} is that the codec adds a residual term for correction. 
	%The correction step addresses the bias problem and the error accumulation problem in image space. 
	Inspired by this, \textbf{we propose to learn the residual term in feature space}~(\cref{fig:inter-frame}(b)) to reduce errors incurred by warping.

    %\uwave{Unfortunately, recent codecs do not have third-party toolkits for us to extract motion vectors and residuals. Thus, we use MPEG-4 part 2 following the works of [33],[34].}

	Recent codecs such as HEVC (H.265) %\mbox{\cite{sullivan2012overview,sze2014high}},
	 and VCC (H.266) share the same motion vector and residual design as MPEG-4 part 2 (H.263) and MPEG-4 part 10 (H.264). %However, the design of motion compensation become more and more sophisticated. %Specifically, all the macro-blocks in H.263 have the same size of $16 \times 16$. H.264 takes a step forward by dividing $16 \times 16$ macroblocks into smaller partitions such as $16 \times 8$, $8\times 8$, and each partition has its own motion vector. Furthermore, H.265 replaces macroblocks with coding tree units which can use larger blocks up to $64 \times 64$. Recently, H.266 introduces block-based affine transform motion compensation and adaptive motion vector resolution, making the motion compensation more accurate.
%Advanced motion compensation is a double-edged sword for the warping-based methods. 
More accurate motion vectors alleviate the error incurred by warping, whereas they also require more time on decoding, making the whole system run slower. No matter how sophisticated the motion compensation is, they still need correction in the compressed domain. It is expected that our method, which does correction in feature space, will still work for these codecs.

\begin{figure}[h]
\begin{adjustwidth}{-\extralength}{0cm}
\begin{center}
   \includegraphics[width=.9\linewidth]{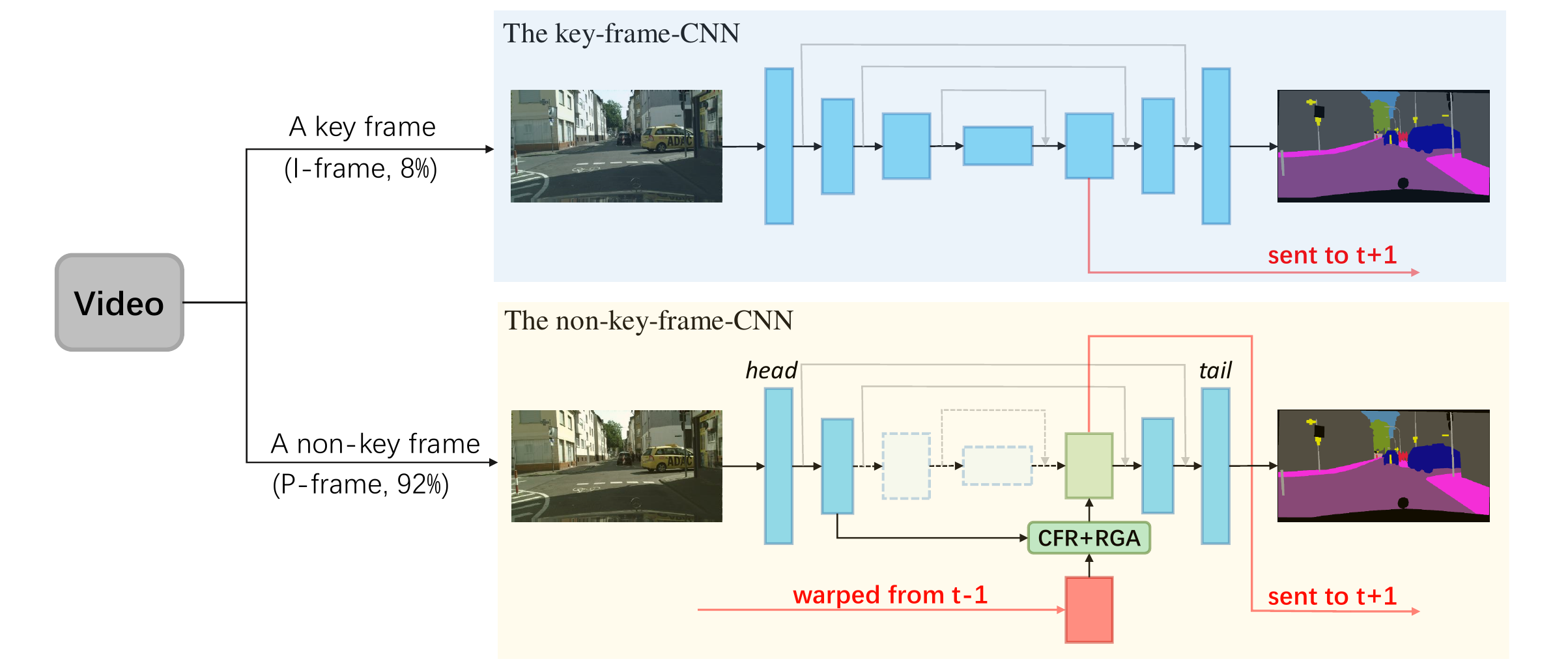}
\end{center}
\end{adjustwidth}
	\vspace{-0.8em}

   \caption{ The framework of \ours. Key frames~are sent to the regular per-frame CNN and non-key frames to the non-key-frame CNN, where the warped context features are corrected.  %Both branches output the result label maps and the interior context feature maps.
   }
\label{fig:wrnet}
\end{figure}
\begin{figure}[htb]
\begin{center}
   \includegraphics[width=0.9\linewidth]{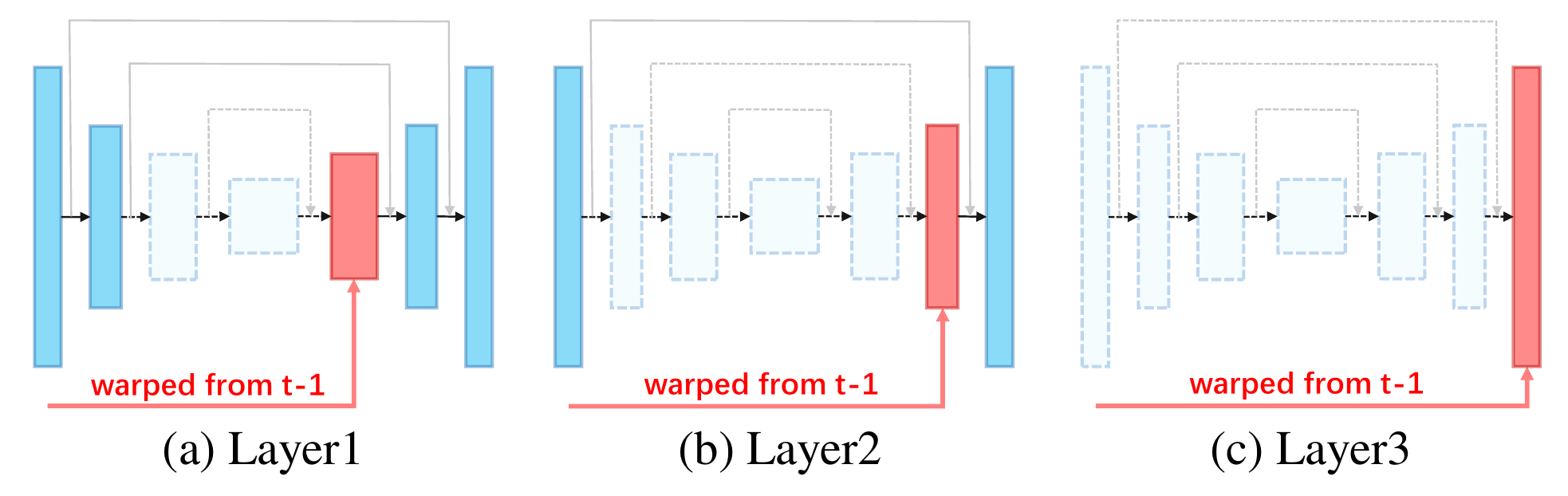}
\end{center}
	\vspace{-0.8em}
   \caption{Feature warping at different layers. The warped layer is in red. The dotted arrows and boxes denote skipped operations.}
\label{fig:multi-layer}
\end{figure}
\begin{figure}[htb]
\begin{center}
   \includegraphics[width=0.8\linewidth]{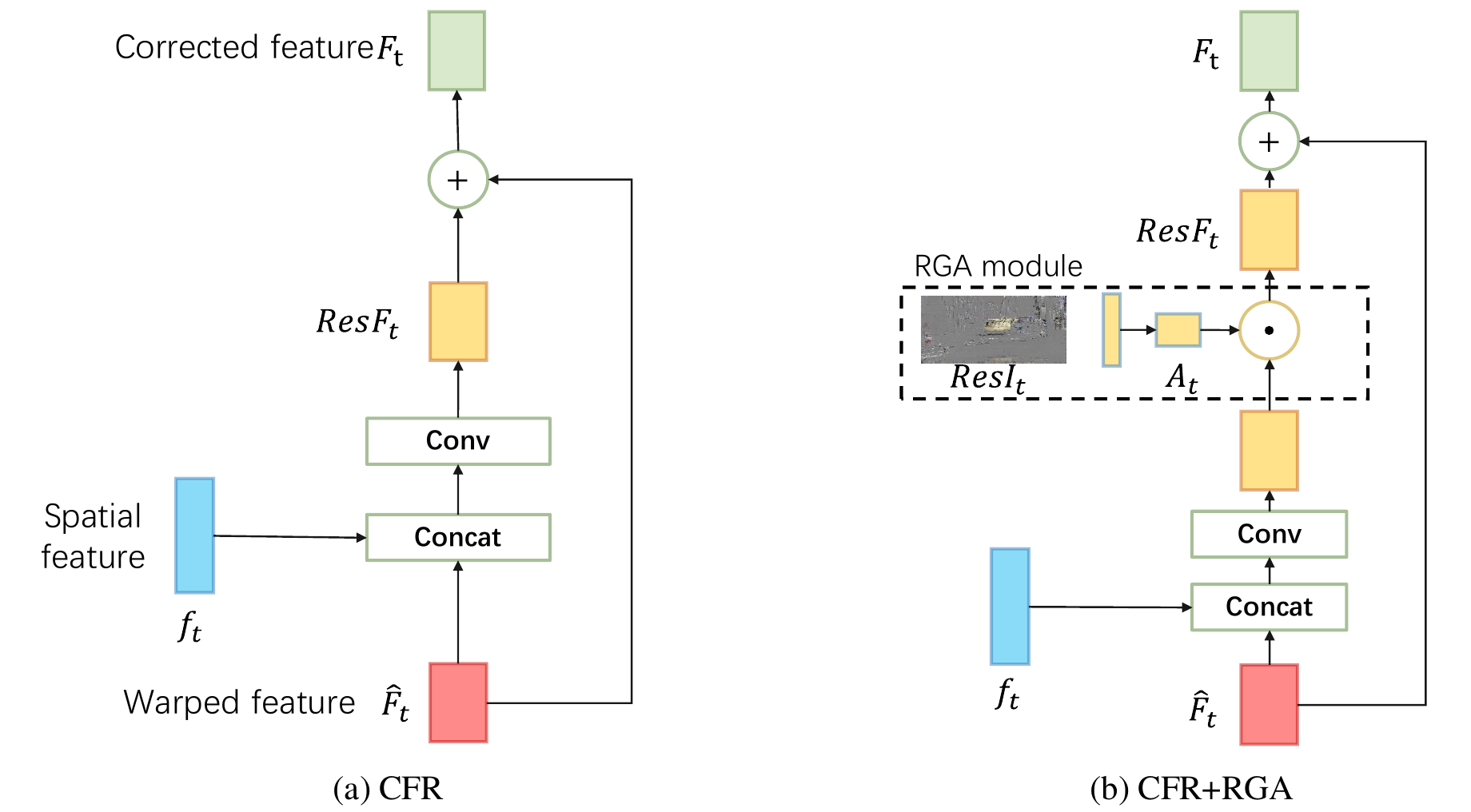}
\end{center}
	\vspace{-0.8em}
   \caption{Details of (a)~CFR and (b)~CFR+RGA. ``$\odot$'': element-wise multiplication; ``$+$'': element-wise addition. }
\label{fig:cfr-rga}
\end{figure}
\section{Tamed Warping Network}
\label{sec:methods}

	% In this section, we introduce the Tamed Warping Network~(TWNet).
	% We first outline the overall framework.
	% Next, we introduce our non-key-frame CNN architecture and the correction modules, i.e., the Context Feature Rectification~(CFR) module as well as the Residual-Guided Attention~(RGA) module. 
	% Finally, we present the implementation details. 

% Overview
%\subsection{Overview}

	Tamed Warping Network~(TWNet) consists of a per-frame CNN, a non-key-frame CNN~(NKFC), and two modules for correction, as illustrated in~\cref{fig:wrnet}. In \ours, video frames are divided into key and non-key frames. Key frames are sent to the per-frame CNN where the context features of a selected interior layer are warped to the next frame. Non-key frames are sent to NKFC and the warped features are corrected by the Context Feature Rectification~(CFR) and Residual-Guided Attention~(RGA) modules. 
	% the Warping-CNN receives the context features from the previous frame, which is fused with currently-extracted spatial feature using the Context Feature Refinement~(CFR) and Residual-Guided Attention~(RGA) modules for correction.
The corrected features are used to make prediction and warped to the next frame. % The components of TWNet are described as follows. %in the following part.

\subsection{Basic Model: The Non-Key-Frame CNN (NKFC)}
\label{sec:warping-cnn}

	% featire warping
	We adopt an encoder-decoder architecture, \eg FPN~\cite{fpn} and U-Net~\cite{unet}, to build our per-frame CNN, and the NKFC, which adopts feature warping, is based on the per-frame CNN.
	%Warping can be applied in feature space~(Fig.~\ref{fig:inter-frame}(b)) since the ``conv'' layers in FCN models preserve positions. 
	Formally, let $t$ be the subscript of the current frame, $t-1$ be that of the previous frame. Given features $\mathcal{F}_{t-1}$, we can first resize the motion vectors  $Mv_{t}$ to $\hat{Mv_{t}}$ to match the size of $\mathcal{F}_{t-1}$, and then predict the features of the current frame $\hat{\mathcal{F}}_{t}$ by 
\begin{equation}
\begin{split}
	\mathcal{\hat{F}}_t = \mathrm{warp}(\mathcal{F}_{t-1}, \hat{Mv_{t}}).
\end{split}
\label{eq:wocfr}			
\end{equation}
The motion vectors we use for warping are readily available from the compressed domain, sparing one from time-consuming motion estimation such as optical flows. Details can be found in \cref{sec:imp-details}.

%\subsection{Layer Selection in the Non-Key-Frame CNN}\label{ssub:layer_selection}
	Note that the CNN layer used for warping can be arbitrarily chosen. In general, if we choose a deeper layer, there will be fewer paired \textit{head} and \textit{tail layers}~(\cref{fig:multi-layer}), since layers in the encoder-decoder architecture are paired by lateral connections. Thus, we will obtain a faster while less accurate model.  For example, if we choose tail2 to do warping, as shown in \mbox{\cref{fig:multi-layer}(b)}, only head1, tail1, and tail2 layers will be working. 
	%In practice, we can adjust this hyperparameter to strike a balance between speed and accuracy.
	Different from previous warping methods, NKFC fuses features from \emph{head} layers to those of \emph{tail} layers to enhance spatial details. Compared to direct lateral connections as FPN~\cite{fpn} and U-Net~\cite{unet}, the spatial features from head layers are first fed into the correction modules as described in~\cref{sec:cfr}.

As for key frame scheduling, we simply regard all \mbox{I-frames} as key frames and P-frames as non-key frames, where I/P-frames are  the concepts in video codecs. %An I-frame~(Intra-coded picture) is stored as a complete image, while a P-frame~(Predicted picture) is stored by the corresponding motion vectors and residual. 
Following~\cite{dmcnet, coviar}, we choose MPEG-4~Part~2~(Simple Profile)~\cite{h264} as the compression standard, where each group of picture contains an I-frame followed by 11 \mbox{P-frames}. %Technical details can be found in Section A of the Supplementary Material.

\subsection{The Correction Stage}
\label{sec:cfr}
	 
	Although NKFC speeds up video segmentation by doing feature warping, errors are inevitably introduced by warping and they will accumulate along successive non-key frames. 	Comparing the pipeline of video codecs with warping-based methods~(\cref{fig:inter-frame}), we found that the main problem of previous methods is the lack of a correction stage. %, which leads to inaccurate predictions and error accumulation. 
	We propose a correction stage %to correct the warped feature by learning the \textit{residual} term in feature space using a context feature refinement~(CFR) module. 
	%$Besides, observing the spatial consistency between image space and feature space, we propose to use the residual map in image space to guide the learning of CFR in an attention manner. %by employing a residual-guided attention~(RGA) module.
	%
	consisting of the following two modules.
	
	% 显示进行 refine
\subsubsection{Context Feature Rectification (CFR)}
	
	 We introduce a lightweight module called CFR to explicitly correct the warped context features considering the following observations. First, the contextual information of the warped features is generally correct, except for the \textit{edges of moving objects}. Second, the low-level features contain the spatial information, such as ``edge'' and ``shape'', which can help to correct the context features. Thus, we make
	CFR take as the input the warped context features $\hat{\mathcal{F}}_{t}$ as well as the spatial features of the current frame $f_t$ and outputs the corrected context features $\mathcal{F}_t$, as shown in \cref{fig:cfr-rga}(a). Specifically, CFR adopts a single-layer network, $\phi_r$, which takes the concatenation $[\hat{\mathcal{F}}_{t}$, $f_t]$ as the input and outputs $Res\mathcal{F}_t$, the residuals in feature space, \ie,
\begin{equation}
\begin{split}
	\mathcal{F}_t &= \hat{\mathcal{F}}_{t} + \phi_{r}([\hat{\mathcal{F}}_{t}, f_t])\\
	&=  \mathrm{warp}(\mathcal{F}_{t-1}, \hat{Mv_{t}}) + Res\mathcal{F}_t.
\end{split}
\label{eq:withcfr}			
\end{equation} 
%
%
	% During the training of CFR, in addition to the commonly used softmax cross entropy loss $\mathcal{L}_{cls}$ for pixel-level classification, we propose to employ an L2 consistency loss $\mathcal{L}_{consist}$ to minimize the distance between the corrected context features and the features extracted by the per-frame CNN.% We add this consistency loss since we expect the output of Warp-FCN is as similar as that of the per-frame FCN. 

\subsubsection{Residual-Guided Attention (RGA)}
%\subsection{Residual-Guided Attention}
	% consistency
	To guide the learning of CFR, we propose the RGA module.  In TWNet, the motion vectors used for feature warping are the same as those used in image warping. % Therefore, the \textit{errors should appear in the same spatial regions} in image space and feature space. 
	Thus, the residual maps in image space $ResI_t$ can be used as prior knowledge to guide the learning of residuals in feature space $Res\mathcal{F}_t$. To this end, %we adopt a lightweight spatial attention module to let the CFR module focus more on the regions with higher responses in $ResI_t$. 
	%In above sections, we apply warping in feature space using the motion vectors extracted from image space. This works since these two spaces are spatially consistent~(mainly because operations in CNN preserve the spatial structure). We claim this consistency also holds for \textit{residuals} in these two spaces, which means high response should appear in the same spatial regions in these two kinds of \textit{residuals}. Based on this, we propose the RGA module to guide the learning of CFR by fully exploiting the prior spatial information contained in residuals from image space.
	%Specifically, 
	we first resize the residual map~$ResI_t$ to the shape of the warped context features. Then, we calculate the spatial attention map~$\mathcal{A}_t$ using a single-layer feed-forward CNN~$\phi_a$ as follows
\begin{equation}
\begin{split}
	\mathcal{A}_t &= \phi_a(ResI_t).
\end{split}
\label{eq:attention}			
\end{equation}
	Finally, we apply spatial attention by performing element-wise multiplication between $\mathcal{A}_t$ and $Res\mathcal{F}_t$. \cref{fig:cfr-rga}(b) illustrates the detailed structure of RGA. After applying this module, %we obtain Eq.~\ref{eq:wrnet}, which gives the whole process of the proposed WRNet for non-key frames.
we formulate the whole process of TWNet for non-key frames as 
\begin{equation}
\begin{split}
	\mathcal{F}_t &= \hat{\mathcal{F}}_{t} + \mathcal{A}_t\odot Res\mathcal{F}_t.
\end{split}
\label{eq:wrnet}			
\end{equation}

\subsection{Training of TWNet}\label{sssec:training}
	The training of TWNet contains two main steps, i.e., the training of the per-frame CNN and the training of NKFC. The training of the per-frame CNN is similar to that of other image segmentation methods, which can be defined by 
\begin{equation}
\begin{split}
	\mathcal{L}_{pf} = \mathcal{L}_{cls} + \lambda_0\cdot\mathcal{L}_{reg},
\end{split}
\label{eq:loss1}			
\end{equation} 
	where  $\mathcal{L}_{cls}$ is the softmax cross entropy loss~ and $\mathcal{L}_{reg}$ is the L2 regularization term. 
%
%	where $\theta_0$ contains trainable parameters in the modified FCN.
	Then, we fix all the parameters in the per-frame model and start to train the modules in NKFC. We employ an additional L2 consistency loss $\mathcal{L}_{consist}$ to minimize the distance between the corrected features and the context features extracted from the per-frame CNN. The object function of this step is 
\begin{equation}
\begin{split}
	%\mathop{\min}~~
	\mathcal{L}_{nkf} = \mathcal{L}_{cls} + \lambda_0\cdot\mathcal{L}_{reg} + \lambda_1\cdot\mathcal{L}_{consist}.
\end{split}
\label{eq:loss2}			
\end{equation} 
%	
	% where $\lambda_0$ and $\lambda_1$ are weights to balance different kinds of losses.

\subsection{Implementation Details}\label{sec:imp-details}
% \section{Motion Extraction and Key Frame Selection} 
In our TWNet, we utilize \textit{motion vectors} as motion maps. Both \textit{motion vectors} and \textit{residuals} are readily available in compressed videos. Thus, it takes no extra time to extract them. As for key frame scheduling, we simply regard all I-frames as key frames and P-frames as non-key frames, where I/P-frames are  the concepts in video compression. An I-frame~(Intra-coded picture) is stored as a complete image, while a P-frame~(Predicted picture) is stored by the corresponding motion vectors and residual. Following previous works of~\cite{dmcnet, coviar}, we choose MPEG-4~Part~2~(Simple Profile)~\cite{h264} as the compression standard, where each group of picture~(GOP) contains an I-frame followed by 11 P-frames.

% We also conduct experiments to show that the refinement modules are able to alleviate error accumulation. 

\begin{table}[htb]
\centering
\caption{Performance comparison of different layers used for feature warping. Feature warping: the layer of feature map used for feature warping; Fine-tuned: whether the second training step is performed to fine-tune the non-key-frame CNN. If not, the parameters of the head and tail layers keep the same as those in the per-frame CNN. If Layer~3 is chosen, there exists no trainable parameters and hence no fine-tuning.}
\label{table:warping-cnn}
\begin{tabular}{l|ccc}
\toprule
Warping Layer\ \  &\ Fine-tuned\ &\ mIoU\ &\ FPS\ \\
\midrule
\midrule
\multirow{2}{*}{Layer~1}&~&67.3&65.5\\
&$\checkmark$&69.6&65.5\\
\midrule
\multirow{2}{*}{Layer~2}&~&65.4&89.8\\
&$\checkmark$&67.8&89.8\\
\midrule
Layer~3&-&63.2&119.7\\
\bottomrule
\end{tabular}
\end{table}	

\begin{table}[htb]
\caption{Validation of $\mathcal{L}_{consist}$. % during the training of CFR. 
$\lambda_1$: the weight of $\mathcal{L}_{consist}$.}
\label{table:cfr}
\centering
\begin{tabular}{l|ccc}
\toprule
Warping Layer\ \ &\ $\lambda_1$ \ & \ mIoU \ & \ FPS \ \\
\midrule
\midrule
\multirow{4}{*}{Layer~1}&0&69.9&63.1\\
&1&70.2&63.1\\
&10&\textbf{70.6}&63.1\\
&20&70.3&63.1\\
\midrule
\multirow{4}{*}{Layer~2}&0&67.6&86.3\\
&1&68.1&86.3\\
&10&\textbf{68.6}&86.3\\
&20&68.3&86.3\\
\bottomrule
\end{tabular}
\end{table}

\begin{table}[tb]
\caption{Effect of each module of TWNet. FT: the fine-tuning of the non-key CNN~(the second training step); CFR: Context Feature Rectification; RGA: Residual-Guided Attention. ``$\checkmark$'' means the model utilizes the corresponding module.  We also show the extra cost of adding our modules.}
\centering
\begin{tabular}{l|cccccc}
\toprule
Warping Layer \ \ & \ FT \ & \ CFR \ & \ RGA \ & \ mIoU \ & \ FPS \ & GFLOPs \\
\midrule
\midrule
\multirow{4}{*}{Layer~1}&&&&67.3&65.5 & 113.28\\
&$\checkmark$&&&69.6&65.5 & +0 \\
&$\checkmark$&$\checkmark$&&70.6&63.1 & +2.42\\
&$\checkmark$&$\checkmark$&$\checkmark$&\textbf{71.6}&61.8 & +0.0012\\
\midrule
\multirow{4}{*}{Layer~2}&&&&65.4&89.8 & 73.00\\
&$\checkmark$&&&67.8&89.8 & +0\\
&$\checkmark$&$\checkmark$&&68.6&86.3 & +2.42\\
&$\checkmark$&$\checkmark$&$\checkmark$&\textbf{69.5}&84.9 & +0.0029\\
\bottomrule
\end{tabular}
\label{table:rga}
\end{table}	

\begin{figure}[tb]
\begin{adjustwidth}{-\extralength}{0cm}
\begin{center}
   	\includegraphics[width=\linewidth]{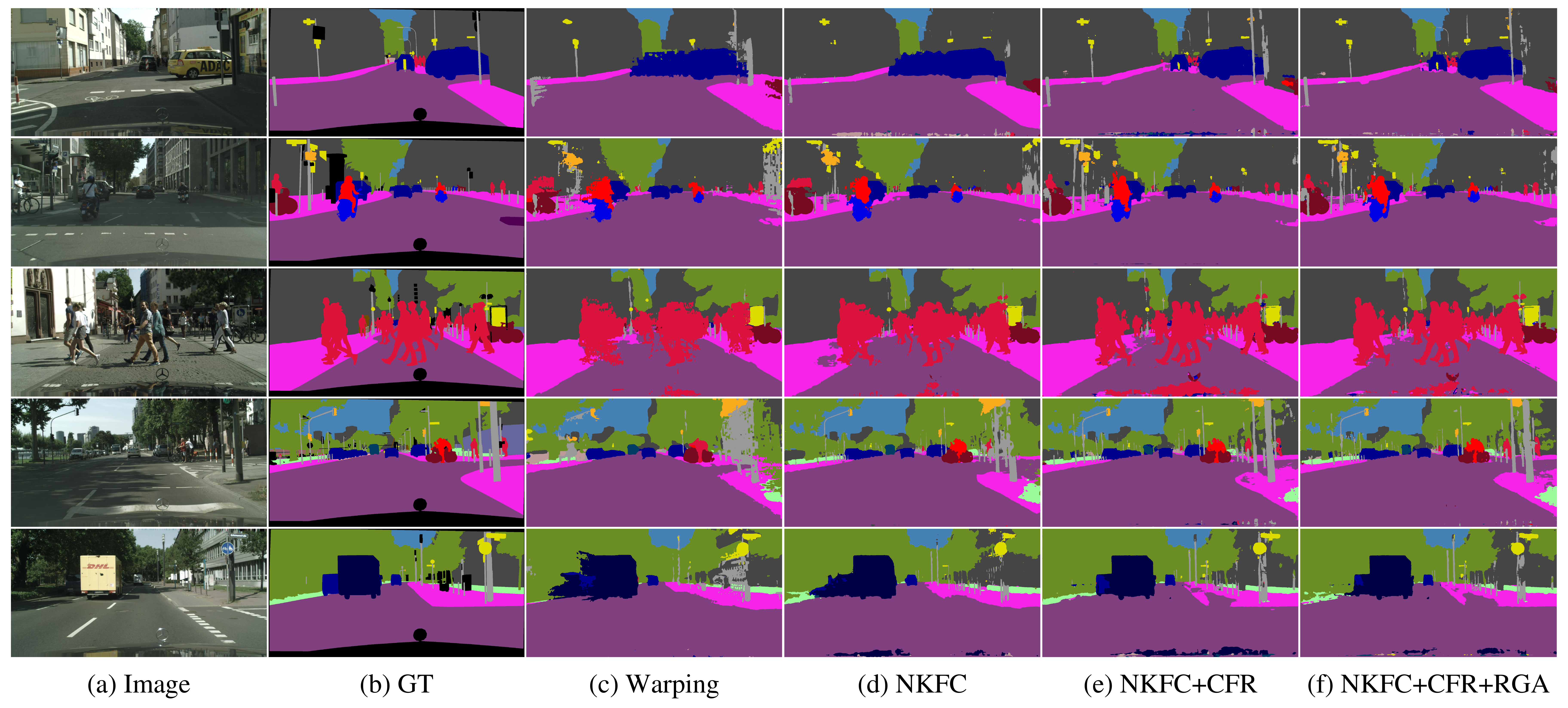}
\end{center}
\end{adjustwidth}
	\vspace{-1em}
   	\caption{Qualitative results on Cityscapes. GT: ground truth; Warping: normal warping; NKFC: the non-key-frame CNN; CFR: context feature rectification; RGA: residual-guided attention.}
\label{fig:qlt}
\end{figure} 

\section{Experiments}
%	In this section, we present the experimental results of TWNet on high-resolution videos. First, we introduce the experimental environment. Then, we perform ablation study to validate the effectiveness of the proposed non-key-frame CNN, CFR, and RGA. Finally, we compare TWNet with the state-of-the-art semantic video segmentation methods.
%
% Experimental Environment	
\subsection{Experimental Setup}

% dataset
%\subsubsection{Datasets} 
	%There are several commonly used datasets (e.g., Cityscapes \cite{cityscapes}, CamVid~\cite{camvid}, COCO-Stuff \cite{coco-stuff}, and ADE20K \cite{ade20k,zhou2019semantic}) for semantic segmentation. 
	We report our major results on the Cityscapes dataset \cite{cityscapes}, which contains 5k images finely annotated with 19 classes. The models are trained on the 2975 training images and evaluated on the 500 validation images. We also obtain results on the 1525 test images, reported by the test server. Each image is the $20^{th}$ frame of a $1024\times2048$ video clip. We also conduct experiments on the CamVid dataset~\cite{camvid}, which can be found in \cref{sec:camvid}.
	
	%We perform the ablation study on the validation set of Cityscapes and compare the results of TWNet with the state-of-the-art methods. Also, we perform experiments on CamVid to demonstrate that our framework is generic to different datasets.

	% \begin{figure}[tb]
	% 	%\centering 
	% 		\begin{subfigure}{0.3\linewidth}
	% 			\centering
	% 			\includegraphics[width=1\linewidth]{figures/major-revision/frankfurt_000000_022254_leftImg8bit.png}\\
	% 			\caption{Raw image} \label{sfig:layer1}
	% 		\end{subfigure}%
	% 		\quad
	% 		\begin{subfigure}{0.3\linewidth}
	% 			\centering
	% 			\includegraphics[width=1\linewidth]{figures/major-revision/rgalv1.png}\\
	% 			\caption{Layer1} \label{sfig:layer1}
	% 		\end{subfigure}%
	% 		\quad
	% 		\begin{subfigure}{0.3\linewidth}
	% 			\centering
	% 			\includegraphics[width=1\linewidth]{figures/major-revision/rgalv2.png}\\
	% 			\caption{Layer2} \label{sfig:layer2}
	% 		\end{subfigure}%
	% 	\centering
	% 	\caption{Visual difference between CFR and CFR+RGA. `Layer1' means we do the Layer1 warping. We can see that the main difference between CFR and CFR+RGA lies in the edges and contours.}
	% 	%\vspace{-0.2cm}
	% 	\label{fig:visual_comparison}
	% \end{figure}

% \subsubsection{Training} 
	% protocal 1. per-frame, 
	The training is divided into two steps, i.e., the training of per-frame CNN and the training of NKFC.
	To train the per-frame model, we use the 2975 fine-annotated training images~(i.e., the $20^{th}$ frames in the video clips). We use MobileNet \cite{mobilenet} pre-trained on ImageNet~\cite{imagenet} as the encoder of the per-frame CNN and three cascaded lateral connections~\cite{fpn} as the decoder. We adopt the Adam optimizer~\cite{adam} to train for $90K$ iterations with the initial learning rate of $1\times10^{-2}$ and a batch size of $8$. We update the pre-trained parameters with a $100$ times smaller learning rate. Weight decay~$\lambda_0$ is set to $1\times10^{-7}$. Training data augmentations include mean extraction, random scaling between $0.5$ to $2.0$, random horizontal flipping, and random cropping to $[800, 800]$. %($[600, 600]$ for CamVid)
	We implement the model using TensorFlow~1.12~\cite{tensorflow} and train it on a GTX~1080~Ti GPU card.
	
	% 2. warp-fcn(no crop)
	After training the per-frame CNN, the parameters in it are fixed, and we start to train NKFC. In each training step of NKFC, we first send a batch of the $19^{th}$ frames into the per-frame CNN to extract their context features. Then, we perform warping and correction for the corresponding $20^{th}$ frames and calculate the loss according to \cref{eq:loss2}. Note that %if the consistency loss $\mathcal{L}_{consist}$ is employed,
	 the $20^{th}$ frames should also be sent to the per-frame model to calculate the context features. Random cropping is not adopted since the warping operation may exceed the cropped boundary. We keep the size of $[1024, 2048]$~%($[720, 960]$ for CamVid)
	 with a batch size of 4. 
	% residual: pre-processing

% \subsubsection{Evaluation} 
At inference time, we conduct all the experiments on video clips at a resolution of $1024\times 2048$. During evaluation, the key frame is uniformly sampled from the $9^{th}$ to the $20^{th}$ frame in the video clip, and the prediction of the $20^{th}$ frame is used for evaluation. No testing augmentation~%(e.g., multi-scale or multi-crop)
 is adopted. The accuracy is measured by mean~Intersection-Over-Union~(mIoU), and the speed is measured by frames per second~(FPS). Our models run on a server with an Intel Core i9-7920X CPU and a single NVIDIA GeForce GTX 1080~Ti GPU card.

\begin{table}[tb]
\vspace{-1.0em}
\caption{IoU improvements of different categories. We choose Layer1 here.}
\centering
\begin{adjustwidth}{-\extralength}{0cm}
\begin{tabular}{l|ccccccc}
\toprule
\small{Method}&\small{object}&\small{human}&\small{vehicle}&\small{nature}&\small{construction}&\small{sky}&\small{flat}\\
\midrule
\midrule
Warping
&43.8&56.7&82.2&86.6&87.1&91.6&96.6\\
NKFC~
&51.2~(+\textbf{7.4})&65.5~(+\textbf{8.8})&84.6~(+2.4)&89.6~(+3.0)&89.1~(+2.0)&94.0~(+2.4)&97.3~(+0.7)\\
NKFC + CFR
&62.2~(+\textbf{18.4})&75.2~(+\textbf{18.5})&89.7~(+\textbf{7.5})&91.3~(+4.7)&90.8~(+3.7)&94.2~(+2.6)&97.9~(+1.3)\\
NKFC + CFR + RGA
&62.2~(+\textbf{18.4})&76.1~(+\textbf{19.4})&90.1~(+\textbf{7.9})&91.1~(4.5)&91.0~(+3.9)&94.2~(+2.6)&98.0~(+1.4)\\
\bottomrule
\end{tabular}
\end{adjustwidth}
\label{table:category}
\end{table}

\subsection{Ablation Study}
	We start building TWNet from the training of the per-frame model. We adopt the commonly used lightweight CNN, MobileNetV1, as the encoder. Our per-frame model achieves the accuracy of $73.6\%$ mIoU at 35.5~FPS. %Results of \textbf{other backbones} and \textbf{lateral connection functions} are listed in the \textbf{supplemental material}.
%
%\begin{table}[t]
%\centering
%\begin{tabular}{lccc}
%\toprule
%Backbone CNN model &mIoU&FPS\\
%\midrule
%\midrule
%MobileNetV1&73.6&30.1\\
%MobileNetV2&73.2&26.7\\
%ResNet-18&72.1&31.2\\
%\bottomrule
%\end{tabular}
%\vspace{0.5em}
%\caption{Performance comparison of our modified per-frame FCN based on different backbones on Cityscapes val set.}
%\label{table:baseline}
%\end{table}	

% Ablation for warping-fcn
\subsubsection{The Non-Key-Frame CNN}
As described in \cref{sec:warping-cnn}, in NKFC, the layer of feature maps can be arbitrarily chosen to balance the accuracy and speed. We choose three layers in the decoder as the context features respectively. Results are summarized in \cref{table:warping-cnn}.

According the experimental results, fine-tuning~(the second training step) can significantly improve the performance. This demonstrates that low-level spatial features are more discriminative in NKFC, possibly due to the fact that the warped context features are less reliable in NKFC, thus the model depends more on low-level spatial features.
%in per-frame FCN, context feature is reliable and the network relies less on these low-level spatial features, while in Warp-FCN, the warped context feature is less reliable, thus the network depends more on  low-level spatial features.

\begin{figure}[t]
    %\centering 
%\begin{adjustwidth}{-\extralength}{0cm}
        \begin{subfigure}{0.45\linewidth}
            \centering
            \includegraphics[width=1\linewidth]{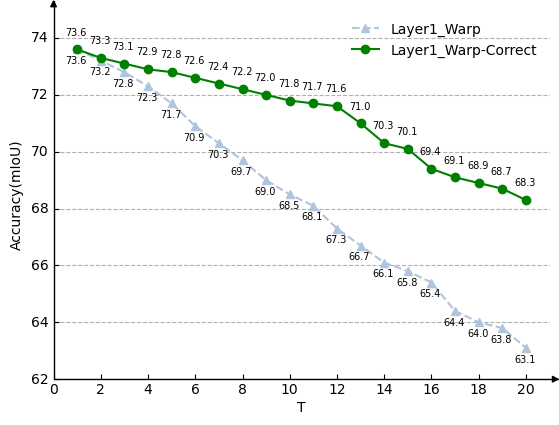}\\
            \caption{Layer1} \label{sfig:20frames-layer1}
        \end{subfigure}%
        \quad
        \begin{subfigure}{0.45\linewidth}
            \centering
            \includegraphics[width=1\linewidth]{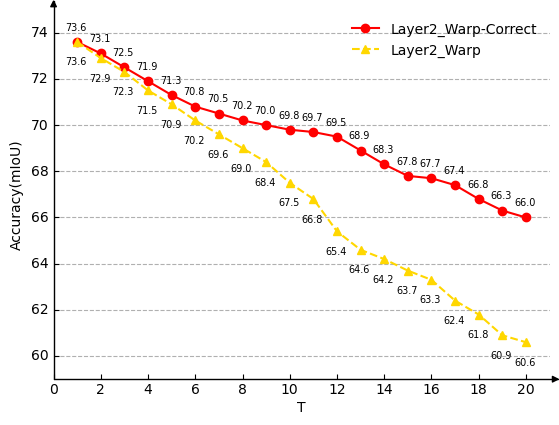}\\
            \caption{Layer2} \label{sfig:20frames-layer2}
        \end{subfigure}%
    \centering
    %\vspace{-0.2cm}
    \label{fig:error-accum}
	%\end{adjustwidth}
   	\caption{Performance degradation of Warp and Warp-Correct. (a): Layer~1 used for feature warping. (b): Layer~2 used for warping. $T$: frame interval between the key frame and the frame to be evaluated. The correction module effectively alleviates the long-term error accumulation problem.}
\end{figure}

% Ablation for refine module
\subsubsection{CFR Module and Consistency Loss}
	% wo kd vs kd
	We propose the CFR module to correct the warped context features. %by learning the residual in feature space. 
	As shown in \cref{table:cfr}, the CFR module is effective and efficient. \cref{table:cfr} also demonstrates the effectiveness of consistency loss, $\mathcal{L}_{consist}$, and the weight term $\lambda_1$, a crucial hyper-parameter for the training of CFR. By default, we set $\lambda_1$ to 10 in the following sections for better performance.
%

% Ablation for residual-attention
\subsubsection{RGA Module}
	We introduce RGA to further exploit the correlation between residuals in image space and feature space. Results are demonstrated in \cref{table:rga}. As expected, the RGA module further improves the performance of TWNet since it guides CFR to pay more attention to error-prone regions. 
	%We do a visual comparison between CFR and CFR+RGA as shown in~\cref{fig:visual_comparison}. We can see that the major differences lie in edges and contours.  
	The qualitative results of TWNet on Cityscapes are shown in \cref{fig:qlt}.

\subsubsection{Category-Level Improvement}
	The IoU improvements for different categories are shown in \cref{table:category}. The IoUs of \textbf{non-rigid objects}~(human, object and vehicle) are improved greatly. The moving of non-rigid objects are hard to predict and thus warping is prone to fail. With our correction, wrong predictions significantly decrease.
%
%
%

% \begin{figure*}[t]
% \begin{center}
%    	\includegraphics[width=0.98\linewidth]{figures/error-accum.pdf}
% \end{center}
% 	\vspace{-1.4em}
% 	% TODO substitute figures
%    	\caption{Performance degradation of Warp and Warp-Correct. (a): Layer~1 used for feature warping. (b): Layer~2 used for warping. $T$: frame interval between the key frame and the frame to be evaluated. The correction module effectively alleviates the long-term error accumulation problem.}
% \label{fig:error-accum}
% \end{figure*}
% robustness -> different interval between
%

\subsubsection{Error Accumulation}
	We also conduct experiments to show that TWNet is able to alleviate the error accumulation problem during consecutive warping. Suppose $T$ denotes the frame-level interval between the initial key frame and the frame to be evaluated. We set $T$ to different values and evaluate the performance of TWNet and NKFC without correction modules. Results in \cref{fig:error-accum} show that the correction modules significantly alleviate the accuracy degradation and improve the robustness of the models. Meanwhile, the employment of CFR and RGA takes little extra time.

% \section{Experiments on Addional Hyperparameters}

%\section{More Analysis of Hyperparameters}

\subsubsection{Choices of Flow Models}

We could use optical flow methods, e.g., FlowNet2~\cite{flownetv2} and PWC-Net~\cite{pwcnet}, as the flow model of our framework. However, the running speeds of these methods are even slower than our per-frame segmentation network (even slower than 20 fps), which means the warping operation will not speed up the inference phase. Also, when we apply these optical flow methods to warping, the segmentation accuracy is similar to our motion-vector version. Thus, we decide to use motion vectors. \cref{table:flow_models} shows the accuracy using different types of flows for warping.

\begin{table}[tb]
\caption{Effect of different flow models.}
\label{table:flow_models}
\centering
\begin{tabular}{l|cccc}
\toprule
Flow model \ \ 		& \ mIoU \	& \ FPS \  \\
\midrule
\midrule
Motion vectors		& 67.3 		& 65.5 \\
FlowNet2			& 67.5 		& 13.2 \\
FlowNet2-s  			& 66.3 		& 26.6 \\
FlowNet2-c  			& 66.6 		& 22.7 \\
PWC-Net   			& 67.0 		& 29.4 \\
\bottomrule
\end{tabular}

\end{table}

\subsubsection{Choices of Backbones} 
We study the genericity of TWNet to different backbones, e.g., MobileNetV1, MobileNetV2 and ResNet-18. 
As described in Section II.A,  we can choose different layers of the decoder for feature warping to build different TWNets. \cref{table:threelayers} shows the required \textit{head} layers for each version.
\begin{table}[tb]
\caption{The required \textit{head} layers for different versions of TWNet. head$n$: the $n^{th}$ \textit{head} layer of the encoder.}
\label{table:threelayers}
	\small
\centering
\begin{tabular}{lcccc}
\toprule
Model name & head1 & head2 & head3\\
\midrule
\midrule
Per-frame& $\checkmark$ & $\checkmark$ & $\checkmark$\\

TWNet-Layer1& $\checkmark$ & $\checkmark$ & \\

TWNet-Layer2& $\checkmark$ &  & \\

TWNet-Layer3&  &  &  \\
\bottomrule
\end{tabular}

%\vspace{0.8em}
\end{table}	
The head layers for different backbones are defined in Table~\ref{table:headlayers}. We follow the notations of the TensorFlow Slim package.

\begin{table}[h]
\caption{\textit{Head} layers for different backbones. We quote the notations from the TensorFlow Slim package.}
\label{table:headlayers}
	\small
\centering
\begin{tabular}{lcccc}
\toprule
Backbone & head1 & head2 & head3\\
\midrule
\midrule
MobileNetV1& $conv2d\_3$ & $conv2d\_5$ & $conv2d\_11$\\

MobileNetV2& $layer\_4$ & $layer\_7$ & $layer\_14$ \\

ResNet-18& $conv2\_2$ & $conv3\_2$ & $conv4\_2$ \\
\bottomrule
\end{tabular}

%\vspace{0.8em}
\end{table}
\begin{table}[tb]
\caption{Performance of TWNet based on different backbone networks. IW: the layer where feature warping is performed. ``None'' means no feature warping.}
\label{table:backbone}
	\small
\centering
\begin{tabular}{llcccc}
\toprule
	Backbone & IW & mIoU & FPS & Speed-up ($\times$)\\
\midrule
\midrule
	\multirow{4}{*}{MobileNetV1} & None & 73.6 & 35.5 & -\\
	&Layer1& 71.6 & 61.8 & 1.74\\
	&Layer2& 69.5 & 84.9 & 2.39\\
	&Layer3& 63.2 & 119.7 & 3.37\\

\midrule
	\multirow{4}{*}{MobileNetV2} & None & 73.2 & 32.3 & -\\
	&Layer1& 71.3 & 59.6 & 1.85\\
	&Layer2& 69.4 & 82.5 & 2.55\\
	&Layer3& 62.4 & 115.8 & 3.59\\

\midrule
	\multirow{4}{*}{ResNet-18} &  None & 71.6 & 36.9 & -\\
	&Layer1& 69.4 & 63.6 & 1.72\\
	&Layer2& 67.7 & 86.8 & 2.35\\
	&Layer3& 61.9 & 120.4 & 3.26\\

\bottomrule
\end{tabular}

\end{table}

The results in \cref{table:backbone} demonstrate that TWNet is generic to backbone networks, and hence can be adapted to various scenarios for different requirements of speed and accuracy.

TWNet is also generic to the choice of backbone %and the choice of flow model 
as discussed in Section B of Supplemental Material.

\begin{table}[t]
%\caption{Comparison of state-of-the-art semantic video segmentation models on Cityscapes. mIoU-pf: accuracy for per-frame model; FPS-pf: speed for per-frame model; mIoU: accuracy for video-based methods; FPS: speed for video-based methods. ``FPS~norm'' is calculated based on the performance of GPU. For previous works, we report the results in \textbf{the most comparable evaluation settings} to ours.}
\caption{Comparison of SOTA models on Cityscapes. \textbf{Terms with ``-pf'': mIoU/FPS for per-frame model}; ``FPS~norm'' is calculated based on the ability of the GPU. All the results only use \emph{train} as the training set. All TWNet models run at a resolution of $1024\times2048$. }
% For previous works, we report the results in \textbf{the most comparable evaluation settings} to ours.
\label{table:cmp}
\centering
\begin{adjustwidth}{-\extralength}{0cm}
\begin{tabular}{l|cccccccc}
\toprule
Model &  \ eval set \ & \ resolution \ & \ mIoU-pf \ & \ mIoU \ & \ FPS-pf \ & \ FPS \ & \ FPS norm \ & GPU\\
\midrule
\midrule
\multicolumn{2}{l}{\textit{Per-frame Models}}\\
ICNet~\cite{icnet}&  val & $1024\times 2048$ & 67.7 & - & 30.3 & - & 49.7 & TITAN X(M)\\
ERFNet~\cite{erfnet}&  test & $1024\times 2048$ & 69.7 & - & 11.2 & - & 18.4 & TITAN X(M)\\
%ICNet& train+val & test & 69.5 & $1024\times 2048$ & 30.3 & 49.7 & TITAN X(M)\\
%BiSeNet& train & val & 69.0 & $768\times 1536$ & 105.8 & 98.8 & TITAN Xp\\
SwiftNetRN-18~\cite{swiftnet}&  val & $1024\times 2048$ & 74.4 & - & 34.0 & - & 34.0 & 1080~Ti\\
%DFANet & train & test & 71.3 & $1024\times 1024$ & 100.0 & 103.0 & TITAN X\\
CAS~\cite{cas}&  val & $1024\times 2048$ & 74.0 & - & 34.2 & - & 48.9 & 1070\\
%CAS & train+coarse & val & 72.5 & $768\times 1536$ & 108 & 127.4 & 1070\\
%Ours~(modified FCN) & train & val & 73.6 & $1024\times 2048$ & 31.3 & 31.3 & 1080~Ti\\
% resolution is doubtful. can be 769*769
Liu et al.\mbox{\cite{liu2020efficient}} &  val & $1024\times 2048$ & 73.8 & - & 20.8 & -  & 20.8 & 1080 Ti \\
TD-PSP18 \mbox{\cite{hu2020temporally}} &  val & $1024\times 2048$ & 76.8 & - & 11.8 & -  & 10.5 & TitanXp \\

\midrule
\midrule
\multicolumn{2}{l}{\textit{Video-based Models}}\\
DFF~\cite{dff}&  val & $512\times 1024$ & 71.1 & 69.2 & 1.52 & 5.6 & 12.8 & Tesla K40\\
DVSNet1~\cite{dvsnet}&  val & $1024 \times 2048$ & 73.5 & 63.2 & 5.6 & 30.4 & 30.4 & 1080 Ti\\
DVSNet2~\cite{dvsnet}&  val & $1024 \times 2048$ & 73.5 & 70.4 & 5.6 & 19.8 & 19.8 & 1080 Ti\\
Prop-mv~\cite{jain2018fast}&  val & $1024 \times 2048$ & 75.2 & 61.7 & 1.3 & 7.6 & 9.6 & Tesla K80\\
Interp-mv~\cite{jain2018fast}&  val & $1024 \times 2048$ & 75.2 & 66.6 & 1.3 & 7.2 & 9.1 & Tesla K80\\
Low-Latency~\cite{low-latency}&  val & $1024\times 2048$ & 80.2 & 75.9 & 2.8 & 8.4 & - & -\\
LMA \mbox{\cite{lma}} &  val & $512\times 1024$ & 72.1 & 73.7 & 99.0 & 86.2  & 67.2 & 2080 Ti \\

\midrule
\multicolumn{2}{l}{\textit{Ours}}\\
TWNet-Layer1&  val & $1024\times 2048$ & 73.6 & 71.6 & 35.5 & 61.8 & 61.8 & \multirow{4}{*}{1080 Ti}\\
TWNet-Layer1&  test & $1024\times 2048$ & 73.1 & 71.2 & 35.5 & 61.8 & 61.8 & \\
TWNet-Layer2&  val & $1024\times 2048$ & 73.6 & 69.5 & 35.5 & 84.9 & 84.9 & \\
TWNet-Layer2&  test & $1024\times 2048$ & 73.1 & 69.0 & 35.5 & 84.9 & 84.9 & \\
\bottomrule
\end{tabular}
\end{adjustwidth}
\end{table}	
\subsection{Comparison with Other Methods}
% Qualitative comparison(image)
	We compare TWNet with other SOTAs in \cref{table:cmp}. Note that the speed~(FPS) values measured in different models are just listed for reference, since the experimental environments may vary a lot. %Besides, some models reshape the input images to a lower resolution and some do not count the time of BN operations~\cite{bn}. 
	All of our models run on the platform with CUDA 9.2, cuDNN 7.3 and TensorFlow 1.12, and we use the \emph{timeline} tool in TensorFlow to measure the speed. Following the recent work of~\cite{swiftnet}, we include the ``FPS norm'' value based on the GPU types of previous methods\footnote{GPU Benchmark:~\url{www.techpowerup.com/gpu-specs}}.
	%
	% relation with image-based methods
	Results demonstrate that TWNet achieves the highest inference speed with comparable accuracy at a resolution of $1024\times 2048$. The accuracy of TWNet decreases more slightly than other video-based methods. % It is worth noting that we adopt the per-frame CNN to simplify our presentation. If a more delicate per-frame model were used, the performance of TWNet would be further improved.
%

	% FAIR!  FAIR!
	% Compare with other methods(big table)

\subsection{Results on the CamVid Dataset}
\label{sec:camvid}
We also conduct experiments on the CamVid dataset, which contains 367, 100 and 233  video clips for training, validation and testing respectively at a resolution of $720\times 960$. We apply the same configurations as those of Cityscapes except for the crop size. As shown in \cref{table:camvid_ablation} and \cref{table:camvid_comparison}, TWNet achieves consistent results on CamVid. 

\begin{table}[h]
\caption{Effect of each module on the CamVid test set. }
\centering
\begin{tabular}{l|ccccc}
\toprule
Warping Layer \ \ & \ FT \ & \ CFR \ & \ RGA \ & \ mIoU \ & \ FPS \ \\
\midrule
\midrule
\multirow{4}{*}{Layer~1}&&&&68.8&183.1\\
&$\checkmark$&&&69.9&183.1\\
&$\checkmark$&$\checkmark$&&71.0&179.8\\
&$\checkmark$&$\checkmark$&$\checkmark$&\textbf{71.5}&175.2\\
\midrule
\multirow{4}{*}{Layer~2}&&&&66.7&252.6\\
&$\checkmark$&&&68.1&252.6\\
&$\checkmark$&$\checkmark$&&69.3&245.8\\
&$\checkmark$&$\checkmark$&$\checkmark$&\textbf{70.0}&240.7\\
\bottomrule
\end{tabular}
\label{table:camvid_ablation}
\end{table}

\begin{table}[h]
\caption{Comparison with others on the CamVid test set. }
\label{table:camvid_comparison}
\centering
\begin{tabular}{l|cccc}
\toprule
Model \ \ 						& \ mIoU-pf \ 	& \ mIoU \	& \ FPS-pf \ & \ FPS \  \\
\midrule
\midrule
DFANet A \cite{dfanet} 		& 64.7 			& - 		& 120		 & - \\
ICNet \cite{icnet} 			& 67.1			& - 		& 27.8 		 & - \\
BiSeNet \cite{bisenet} 		& 68.7 			& - 		& -			 & - \\
Liu et al. \cite{liu2020efficient}  & 78.2 			& - 		& 27.8		 & - \\
TD-PSP18 \cite{hu2020temporally}  & 72.6 			& - 		& 25		 & - \\
Prov-mv \cite{jain2018fast} 	& 68.6 			& 63.4 		& 3.8		 & 21.4 \\
Interp-mv \cite{jain2018fast}  & 68.6 			& 67.3 		& 3.8		 & 19.1 \\
% DFANet A [2] 		& 64.7 			& - 		& 120		 & - \\
% ICNet [15] 			& 67.1			& - 		& 27.8 		 & - \\
% BiSeNet [3] 		& 68.7 			& - 		& -			 & - \\
% Liu et al. [16]  & 78.2 			& - 		& 27.8		 & - \\
% TD-PSP18 [18]  & 72.6 			& - 		& 25		 & - \\
% Prov-mv [20] 	& 68.6 			& 63.4 		& 3.8		 & 21.4 \\
% Interp-mv [20]  & 68.6 			& 67.3 		& 3.8		 & 19.1 \\

\midrule
TWNet-Layer1 					& 73.5			& 71.5		& 103.5 	 & 175.2 \\
TWNet-Layer2 					& 73.5 			& 70.0 		& 103.5		 & 240.7 \\
\bottomrule
\end{tabular}

\end{table}	
\section{Conclusion}
	We present a novel framework TWNet for fast high-resolution semantic video segmentation. % Our TWNet efficiently improves the performance and robustness of warping-based segmentation methods. 
	Specifically, we use warping and employ NKFC for acceleration. 
In order to alleviate the errors caused by feature warping, we propose two efficient modules, namely CFR and RGA, to correct the warped features by learning the feature-space residuals. Experimental results demonstrate that our method is much more robust than previous warping-based approaches while keeps the high speed.

\vspace{6pt} 

\begin{adjustwidth}{-\extralength}{0cm}

%\newpage
\reftitle{References}


%\bibliography{your_external_BibTeX_file}
\bibliography{bibli}

\PublishersNote{}
\end{adjustwidth}
\end{document}